\pdfoutput=1

\documentclass[11pt]{article}

\usepackage[final]{acl}
\usepackage{times}
\usepackage{latexsym}
\usepackage{enumitem}
\usepackage{multirow}
\usepackage{booktabs}
\usepackage{listings}
\usepackage{amsfonts}
\usepackage{stfloats, caption, subcaption}
\usepackage[T1]{fontenc}

\usepackage[utf8]{inputenc}

\usepackage{microtype}

\usepackage{inconsolata}

\usepackage{graphicx}
\usepackage{xcolor}
\usepackage{colortbl}
\usepackage{float}
\usepackage{graphicx,multirow}
\usepackage{pifont}
\usepackage{float}
\usepackage{multicol}
\usepackage{rotating}
\usepackage{enumitem}
\usepackage{tabularx}
\usepackage{nicematrix}
\usepackage{pdfpages}

\newcommand{\tightparagraph}[1]{\noindent\textbf{#1}}

\setlist[itemize]{noitemsep}
\title{Your Students Don't Use LLMs Like You Wish They Did}

\author{
  \textbf{Sebastian Kobler},
  \textbf{Matthew Clemson},
  \textbf{Angela Sun},
  \textbf{Jonathan K. Kummerfeld}
\\
  The University of Sydney
\\
  \small{\texttt{\{skob7546, matthew.clemson, angela.sun, jonathan.kummerfeld\}@sydney.edu.au}}
}

\begin{document}
\maketitle
\begin{abstract}
Educational NLP systems are typically evaluated using engagement metrics and satisfaction surveys, which are at best a proxy for meeting pedagogical goals. We introduce six computational metrics for automated evaluation of pedagogical alignment in student-AI dialogue. We validate our metrics through analysis of 12,650 messages across 500 conversations from four courses. Using our metrics, we identify a fundamental misalignment: educators design conversational tutors for sustained learning dialogue, but students mainly use them for answer-extraction. Deployment context is the strongest predictor of usage patterns, outweighing student preference or system design: when AI tools are optional, usage concentrates around deadlines; when integrated into course structure, students ask for solutions to verbatim assignment questions. Whole-dialogue evaluation misses these turn-by-turn patterns. Our metrics will enable researchers building educational dialogue systems to measure whether they are achieving their pedagogical goals.
\end{abstract}

\section{Introduction}

Looking at 10 recent ACL papers that tested LLM educational systems with real students, 9 relied on satisfaction surveys and self-reported learning gains, without verifying whether students actually used the system as intended. This mismatch is an issue because extensive educational psychology research has shown that students are systematically poor judges of their own learning, suffering from metacognitive biases like the ``illusion of fluency'' that cause them to mistake ease of interaction for depth of understanding \cite{koriat2005illusions}.

This paper introduces a computational framework for measuring what actually happens when students interact with educational AI systems. We introduce six metrics that capture distinct dimensions of behaviour: conversational engagement, learning orientation, scaffolding resistance, assignment dependency, crisis-mode behaviour, and usage concentration. These metrics provide a novel evaluation infrastructure for validating and analysing educational systems against their intended pedagogical function.

Through analysis of 12,650 messages across 500 student-AI conversations from four courses, we uncover a fundamental misalignment between pedagogical design and student usage patterns. When AI tools are optional, usage concentrates around exam deadlines as crisis management, with 59\% of all semester interactions occurring during a single exam week. When integrated into course structure, students frequently paste assignment questions directly to obtain solutions. These answer-seeking behaviours are masked in whole-dialogue evaluation, but clear in our turn-by-turn analysis. Deployment context shows stronger association with these problematic usage patterns than student preference or system design.

Our framework brings established educational data mining insights into computational dialogue assessment, enabling NLP researchers to move beyond satisfaction surveys toward evidence-based evaluation. We demonstrate that our metrics expose pedagogical failures invisible to standard engagement and satisfaction measures. Crucially, while student preferences for efficiency remain constant, deployment context determines whether these preferences lead to productive engagement or system gaming. External validation on RECIPE4U, a dataset with per-response satisfaction ratings, confirms that satisfaction does not correlate with pedagogical alignment. Our metrics will enable NLP researchers to develop and analyse systems that are truly educational, not just fun to engage with.

\section{Related Work}
Three key issues create the conditions for pedagogical misalignment: (1) well-documented but ignored patterns of problematic student behaviour, (2) metacognitive limitations that make self-assessment unreliable, and (3) flaws in how educational NLP systems are evaluated.

\subsection{Documented but Ignored Student Behaviours}

\tightparagraph{Systems assume dialogue, students seek answers.} Even at the system level, maintaining Socratic dialogue proves difficult: \citet{kumar2024socratic} find that LLMs tasked with generating Socratic questions frequently produce outputs that directly reveal solutions, yet evaluate their fix using automatic generation metrics rather than verifying pedagogical outcomes with real students. This gap between intended pedagogy and actual function persists at the deployment level, where usage prioritises answer-seeking over exploratory learning.

\tightparagraph{Controlled studies miss crisis behaviour.} Researchers typically recruit motivated volunteers rather than exhausted students facing deadlines. This selection bias artificially validates pedagogical assumptions and misses crisis-driven usage patterns. While \citet{xu2024promises} note that model efficacy "diminishes with more complex teaching practices" in real classrooms, no prior work has examined how deployment strategies affect usage patterns.

\tightparagraph{Gaming behaviours are well-documented but rarely measured in NLP systems.} Learning analytics research has extensively documented the behaviours we observe. \citet{vanacore2024effect} found that 94.18\% of students engage in "gaming the system'' behaviours, with \citet{baker2008developing} establishing a framework for understanding gaming as exploiting properties of the system rather than learning.  \citet{chiang2024large} document students "manipulating the LLM to output specific strings" for high scores without meeting rubrics, yet such gaming behaviours are rarely measured.

\tightparagraph{Students concentrate usage around deadlines.} \citet{yang2020prediction} identified procrastination patterns with submission rates accelerating near deadlines, while \citet{ferguson2016consistent} found that MOOC engagement patterns consistently diverge from designers' expectations. These studies explain why optional AI tools become crisis management systems.

\tightparagraph{Scaffolding resistance reflects unproductive help-seeking.} \citet{inproceedings} found that 72\% of help-seeking in intelligent tutoring systems represents unproductive gaming, consistent with their earlier findings \cite{article}. \citet{murray2005effects} demonstrated that minimal friction significantly reduces unproductive help-seeking, supporting our finding that deployment context shapes behaviour more than student preferences.

\tightparagraph{NLP evaluation ignores established evidence of student gaming.} While the gaming behaviours and temporal patterns above come from educational data mining, they have not been systematically incorporated into NLP evaluation of educational dialogue systems. Similarly, HCI research has examined student-AI interaction, revealing over-reliance and direct question copy-pasting in LLM-based assistants \citep{10.1145/3613904.3642773}, and exploring conversational agents for educational feedback \citep{10.1145/3334480.3382805}. However, this prior work focuses on system design and user studies rather than computational metrics for detecting pedagogically misaligned behaviours at scale. Our framework addresses this gap.

\subsection{The Metacognition Crisis}

The "illusion of fluency" is just one aspect of systematic metacognitive failures that undermine self-directed learning with AI. The Dunning-Kruger effect manifests strongly in educational contexts. \citet{kruger1999unskilled} found bottom-quartile students estimated their performance at the 62nd percentile, a 50-point overestimation. This "dual burden" means students least capable of learning effectively are most confident in their approaches.

Students systematically misjudge effective learning. \citet{bjork2011making} demonstrate that students confuse retrieval strength with storage strength, choosing comfortable but ineffective methods. \citet{koriat2005illusions} found strong overconfidence effects when material feels fluent. These findings, including consistently poor self-assessment accuracy \citep{handel2016unskilled, metcalfe2009metacognitive}, explain why features that increase satisfaction ratings may undermine learning, since fluency feels like understanding and students reward the very features that reduce productive struggle.

This creates the "satisfaction-effectiveness inversion": features that make students rate AI tutors highly may undermine pedagogical value. Smooth conversational flow prevents productive struggle. Immediate answers short-circuit reflection. High availability enables procrastination rather than sustained engagement.

\subsection{Emerging Recognition of Evaluation Failures}

\tightparagraph{Technical metrics mask pedagogical failure.} Standard NLP evaluation metrics fail to capture pedagogical effectiveness \cite{tack2023bea}. \citet{sonkar2024pedagogical} acknowledge that supervised fine-tuning "doesn't explicitly favor pedagogically effective responses," highlighting challenges in aligning systems with pedagogical goals.

\tightparagraph{High-performing learner bias reinforces gaps.} \citet{schleifer2024anna} document that LLM embeddings can identify correct student responses but cannot distinguish between different error patterns, meaning systems designed for equitable support fail to detect and help struggling learners, actually reinforcing achievement gaps. While LLMs will improve at detecting errors (a technical limitation), the fundamental bias persists: students who most need help are least equipped to extract it from conversational interfaces \cite{10.1007/3-540-45108-0_33}, a structural issue in how these systems reward academic communication skills.

Recent work acknowledges the evaluation gaps our framework addresses. Our emphasis on multi-dimensional metrics aligns with \citet{rachatasumrit2024beyond}, who call for interpretable, meaningful metrics over prediction accuracy. \citet{zambrano2024says} support our use of multiple complementary metrics by demonstrating that self-report and observational measures produce substantially different results. This disconnect reflects broader patterns in educational technology where system features designed to help can inadvertently harm learning. The 'assistance dilemma' described by \cite{koedinger_exploring_2007} demonstrates that too much help prevents productive struggle, while too little causes frustration. This finding directly parallels our discovery that answer-seeking interactions, while satisfying to students, undermine pedagogical goals.

Further, \cite{vanzo2025gpt4} celebrate GPT-4 as a homework tutor based on student satisfaction and engagement metrics, with 91\% of students wanting to continue using the system. However, their own analysis reveals that engagement, not the treatment condition, predicted learning gains, and improvements were limited to objective-type exercises where answer extraction would be most effective. Critically, their engagement metrics (word count and message frequency) cannot distinguish between productive learning dialogue and answer-extraction behaviours, a distinction our framework makes through turn-by-turn analysis and multiple convergent metrics. Without such granular analysis, what appears as beneficial engagement may actually represent the very gaming behaviours that undermine learning. \citet{berman2024scoping} explicitly distinguish between usability and effectiveness evaluation, noting most AI tools are assessed on ease of use rather than goal achievement, a distinction central to our framework. 

Recent work has proposed complementary evaluation approaches. \citet{maurya-etal-2025-unifying} introduce a taxonomy of eight pedagogical dimensions for assessing AI tutor response quality, while \citet{oliveira2025drive} propose the DRIVE framework for evaluating student learning through GenAI interactions, finding that assessment design shapes whether students use AI for idea development or passive task delegation. Both share our concern with moving beyond satisfaction-based evaluation but differ in focus: \citet{maurya-etal-2025-unifying} evaluate the tutor's quality per-response, \citet{oliveira2025drive} assess learning outcomes through interaction analysis. Our work provides a complementary layer: automated metrics for detecting student behavioural patterns such as crisis-driven usage and scaffolding resistance. These patterns emerge only when analysing sequences of interactions over time, making them invisible to per-response evaluation and difficult to capture through outcome-focused assessment alone.

\section{Behaviour Metrics}

We developed a computational framework to detect and quantify student usage behaviours in educational AI interactions. Our approach uses six novel metrics, validated against human expert judgement (Section~\ref{sec:validation}), enabling scalable detection of usage patterns that diverge from pedagogical intent.

\subsection{Metrics Framework}
We introduce six metrics designed to capture behaviours relevant to pedagogical alignment in student-AI dialogue. To our knowledge, these are the first computational metrics specifically designed for evaluation of such behaviours at scale. We implemented two analysis approaches: \textbf{turn-by-turn analysis} evaluates each student-AI exchange independently, capturing granular behavioural patterns like immediate answer-seeking and scaffolding resistance; \textbf{whole dialogue analysis} processes entire conversations as single units, identifying broader patterns like overall engagement trajectory and assignment characteristics.

Four metrics employ large language models for analysis: LOI, SRS, ADR, and CMI. These LLM-based metrics use zero-shot prompting to measure behavioural indicators, with complete prompts provided in Appendix~\ref{sec:appendix}. All metric weights were set based on the authors' teaching experience and pedagogical judgment, without data-driven tuning. For all metrics, the LLM assigns continuous 0--1 scores rather than binary classifications, allowing proportional values for ambiguous cases (see Appendix~\ref{app:classification_examples} for examples).

\subsubsection{Conversational Engagement Score (CES)}
Drawing on dialogue analysis literature on productive educational discourse \cite{Chi02102014}, CES distinguishes genuine dialogue from transactional information-seeking through a weighted average of four dimensions: log-normalised turn count (sustained interaction), follow-up rate (responses that build on AI answers rather than starting new topics), context reference rate (callbacks to earlier discussion), and acknowledgement rate (engagement markers like ``I see'' or ``that makes sense''). For LLM-analysed components (follow-up, context reference, acknowledgement), we apply binary classification to each student turn, with the final rate calculated as the fraction of messages labeled as positive. Values range 0--1, with higher scores indicating more conversational engagement. See Appendix~\ref{app:CES_details} for implementation details.

\subsubsection{Learning Orientation Index (LOI)}
Adapting Bloom's Taxonomy \cite{bloom1956taxonomy} and help-seeking research \cite{inproceedings}, LOI measures the balance between exploratory learning and solution-seeking through LLM classification. Each student message receives a score (0.0–1.0) distinguishing exploratory markers (process-focused questions, conceptual connections, hypothetical scenarios) from solution-seeking markers (direct answer requests, template seeking, results-focused queries). The index represents the proportion of exploratory interactions:
\begin{equation*}
\text{LOI} = \frac{\sum_{i} \text{exploratory\_weight}_i}{\sum_{i} \text{all\_weights}_i}
\end{equation*}
Values approaching 1 indicate learning-oriented behaviour; values near 0 indicate answer-extraction patterns\footnote{We distinguish ``Answer Seeking'' (LOI category) from ``Answer Extraction'' (behavioural pattern).}. See Appendix~\ref{app:LOI_details} for classification criteria.

\subsubsection{Scaffolding Resistance Score (SRS)}
Building on gaming detection research \cite{baker2008developing}, SRS quantifies student rejection of pedagogical guidance through a three-step process. First, it identifies AI scaffolding attempts (hints, leading questions, Socratic method). Second, it classifies student responses as accepting (engages with guidance), resisting (requests direct answer), or bypassing (ignores guidance). Finally, it calculates resistance proportion:
\begin{equation*}
\text{SRS} = \frac{\text{resist\_count} + 0.5 \times \text{bypass\_count}}{\text{total\_scaffolding\_attempts}}
\end{equation*}
Higher scores indicate greater resistance to pedagogical scaffolding, suggesting preference for direct answers over guided learning. See Appendix~\ref{app:SRS_details} for detection criteria.

\subsubsection{Assignment Dependency Ratio (ADR)}
Combining structural text analysis with teaching observation of homework-driven query patterns, ADR detects assignment-driven usage through parallel rule-based and LLM-based analysis. Rule-based detection identifies structural markers (numbered questions, academic imperatives). LLM analysis evaluates conversational patterns (topic jumping, problem set behaviour). Both methods produce scores from 0--1:
\begin{align*}
\text{ADR}_{\text{rule}} &= \frac{\text{detected\_markers}}{\text{total\_messages}}\\
\text{ADR}_{\text{llm}} &= \text{assignment\_probability}
\end{align*}
Higher values indicate greater likelihood of assignment-related usage. See Appendix~\ref{app:adr_implementation} for implementation details.

\subsubsection{Crisis Mode Indicator (CMI)}
Adapting procrastination research \cite{yang2020prediction}, CMI detects behavioural shifts during high-pressure periods through within-student temporal analysis. It compares peak usage against baseline behaviour using five weighted indicators: panic indicators (urgency language like ``desperate'' or ``exam tomorrow''), query directness shift (more blunt requests), late-night usage (activity outside normal hours), single-exchange ratio (one-and-done queries), and engagement decrease (shorter responses, fewer follow-ups). Values range 0--1, with higher scores indicating crisis-driven usage patterns. See Appendix~\ref{app:CMI_details} for baseline establishment and shift detection methods.

\subsubsection{Usage Concentration Index (UCI)}
Applying the Gini coefficient from economics to temporal usage patterns, UCI measures the distribution of platform usage across a semester through three components: Gini coefficient (inequality in daily usage distribution), normalised peak-to-average ratio (intensity of usage spikes), and temporal clustering (consecutive high-usage periods). Values approaching 1 indicate highly concentrated crisis-driven usage; lower values suggest distributed engagement throughout the semester. See Appendix~\ref{app:UCI_details} for complete formulas.

\begin{table*}
\centering
\small
\setlength{\tabcolsep}{4.75pt}
\begin{tabular}{@{}lcccccccccccccccc@{}}
\toprule
 & & \multicolumn{3}{c}{\textbf{LOI}} & & \multicolumn{3}{c}{\textbf{CES}} & & \multicolumn{3}{c}{\textbf{SRS}} & & \multicolumn{3}{c}{\textbf{ADR}} \\
\cmidrule(lr){3-5} \cmidrule(lr){7-9} \cmidrule(lr){11-13} \cmidrule(lr){15-17}
& & \textbf{r} & \textbf{Exact} & \textbf{±1} & & \textbf{r} & \textbf{Exact} & \textbf{±1} & & \textbf{r} & \textbf{Exact} & \textbf{±1} & & \textbf{r} & \textbf{Exact} & \textbf{±1} \\
\midrule
GPT 4.1-mini Turn & & 0.62 & 66\% & 97\% & & 0.42 & 50\% & 95\% & & 0.64 & 56\% & 88\% & & -- & -- & -- \\
GPT 4.1-mini Whole & & 0.33 & 16\% & 44\% & & 0.21 & 10\% & 58\% & & 0.25 & 39\% & 75\% & & 0.22 & 48\% & 65\% \\
GPT 5 Turn & & \textbf{0.72} & 72\% & 99\% & & \textbf{0.59} & 44\% & 92\% & & \textbf{0.67} & 63\% & 91\% & & -- & -- & -- \\
GPT 5 Whole & & 0.47 & 19\% & 63\% & & 0.46 & 34\% & 81\% & & 0.49 & 47\% & 79\% & & 0.31 & 38\% & 54\% \\
Rule-based & & -- & -- & -- & & -- & -- & -- & & -- & -- & -- & & 0.35 & 82\% & 85\% \\
\midrule
Human-Human$^\dagger$ & & 0.58 & 59\% & 100\% & & 0.67 & 55\% & 88\% & & 0.64 & 60\% & 85\% & & 0.65 & 75\% & 82\% \\
\bottomrule
\end{tabular}
\caption{Model-human and human-human agreement.
r = Pearson correlation (top 5 rows), and weighted Cohen's kappa (bottom row).
±1 = off-by-one accuracy.
$\dagger$ row is based on 100 conversations, while the others are based on 248.
GPT-5 turn-by-turn achieves highest correlations and consistently outperforms whole-dialogue analysis.}
\label{tab:model_agreement_inter-rater}
\end{table*}

\subsection{Design Rationale}

Our metrics rely on heuristic weighting and zero-shot LLM classification. We explain three key design choices below.

\paragraph{Weighting.} All component weights reflect pedagogical priorities rather than data-driven optimisation, as no ground-truth pedagogical outcome data exists against which to tune. Within each metric, the highest-weighted components are those most directly observable and discriminating. For CES, turn count (0.40) receives the highest weight as the strongest signal distinguishing dialogue from transaction, followed by follow-up rate (0.25), context reference (0.20), and acknowledgement (0.15). For CMI, panic indicators (0.30) and query directness shift (0.25) are weighted highest as the most direct behavioural signals of crisis-mode usage, with late-night usage (0.20), single-exchange ratio (0.15), and engagement decrease (0.10) as corroborating indicators.

For ADR, copy-paste indicators (0.40) receive the highest weight as the strongest signal of assignment dependency, followed by problem set behaviour (0.30), answer-seeking focus (0.20), and urgency signals (0.10). For SRS, response weights follow a graduated scale: resisting (1.0) represents clear rejection of scaffolding, bypassing (0.5) reflects passive avoidance, and mixed (0.25) indicates partial engagement.

\paragraph{Continuous scoring.} We chose continuous 0--1 scores over binary classification because pedagogical behaviours exist on a spectrum. A student may shift from exploratory to answer-seeking within a single conversation, or partially engage with scaffolding before requesting a direct answer. Binary labels would obscure these within-conversation dynamics that are central to our analysis. Appendix Table \ref{tab:ambiguous_examples} illustrates how continuous scoring captures these ambiguous cases.

\paragraph{Zero-shot classification.} We use zero-shot prompting without fine-tuning or few-shot examples for all LLM-based metrics. This choice prioritises generalisability across educational domains: fine-tuned classifiers would require labelled training data for each new deployment context, undermining the framework's utility as a general evaluation tool. Our validation (Table \ref{tab:model_agreement_inter-rater}) demonstrates that zero-shot GPT-5 turn-by-turn analysis achieves correlations of 0.59--0.72 with human judgement, approaching inter-rater agreement levels (0.58--0.67 weighted kappa), suggesting that the classification criteria in our prompts are sufficiently well-specified for reliable automated evaluation.

\section{Experiments}
\subsection{Dataset Composition}
We analysed 500 student-AI conversations from five distinct educational datasets spanning three disciplines and two interaction paradigms (see Appendix~\ref{app:sampling} for sampling strategy).

\paragraph{Pedagogical Support Tool Datasets:} We have data from three courses where students interacted with optional AI tools designed to scaffold learning through guided questioning and progressive hints: \textbf{DrMattTabolism}, deployed in Week 3 of a second-year biochemistry course covering metabolism modules (Weeks 1-6); \textbf{DrNucleicAlice}, deployed later in the same course covering molecular biology modules (Weeks 7-12)\footnote{See Appendix section \ref{app:biolearn_prompts} for differences in context prompts between the DrMattTabolism and DrNucleicAlice chat bots.}; \textbf{MEDS2004}, an optional tool in a second-year medical sciences course that provided choreographed practice with past exam questions, following a strict question-answer-feedback sequence; and \textbf{OLiMent} \cite{10.1007/978-3-031-98420-4_10}, used in an introductory data science course for self-reflection on progress.

While all tools were built on similar platforms with pedagogical constraints, their implementations varied. For example, DrMattTabolism employed a flexible questioning approach that would provide answers when pressed but consistently included prompting questions, resulting in shorter conversations (9.0 messages average); DrNucleicAlice enforced stricter Socratic dialogue requiring the AI to respond only with guiding questions until students demonstrated understanding, with scaffolding through Bloom's Taxonomy \cite{bloom1956taxonomy}, leading to longer interactions (14.5 messages average) as students were challenged to engage more deeply before receiving answers.

\paragraph{Unrestricted AI Dataset:} The \textbf{StudyChat} (SC) dataset \cite{mcnichols2025studychat} from an upper-level computer science course where students were encouraged to use an AI assistant throughout the semester for all coursework, represents a contrasting deployment paradigm where AI interaction was integrated into weekly assignments.
\paragraph{}
All datasets represent complete academic semesters (DrNucleicAlice was introduced slightly later in the semester) with student identifiers removed. The original data collection obtained consent from students for research purposes including dialogue analysis. Only StudyChat is publicly available; the remaining datasets cannot be released due to ethics restrictions and privacy requirements.

\subsection{Model Configuration and Validation}
\label{sec:validation}

\paragraph{Human Agreement}
To validate our metrics, we manually labeled 248 conversations (approximately 50 per dataset, with 2 held-out due to non-English text).
The primary rater evaluated all 248 conversations while the second rater independently assessed 100 overlapping conversations to establish inter-rater reliability. The annotators bring complementary pedagogical backgrounds: one is an assistant professor in computer science with 4 years experience as the instructor for university courses; the other is a graduate student with four years of experience as a teaching assistant at university and secondary school levels. Human evaluators rated using a 1-5 scale for continuous metrics (CES, SRS, ADR) and using categories for LOI (Answer-seeking, Mixed, Exploratory).
We found high inter-rater agreement (Table~\ref{tab:model_agreement_inter-rater}, bottom row): weighted Cohen's kappa (with quadratic weights) ranged from 0.58 for LOI to 0.67 for CES, with exact match rates of 55-75\% and within-one-level agreement of 82-100\%. These results confirm both the reliability of our annotation protocol and establish empirical baselines for evaluating model performance.
We use all 248 labeled samples to evaluate the models.

\paragraph{LLM Selection}
We compared two models to evaluate metric reliability and cost-effectiveness. GPT-4.1-mini offered cost-effective processing whereas GPT-5 provided enhanced reasoning capabilities at higher cost with reasoning effort set at the default medium setting. All LLM-based metrics employed zero-shot prompting without examples or fine-tuning, with complete prompts provided in the respective appendix sections. Cost analysis is available in Appendix~\ref{app:cost-analysis}.

\paragraph{Model Validation}
ADR showed the weakest computational-human alignment. LLM-based approaches averaged 3.00 compared to human ratings of 1.49, an overestimation of +1.51 points. LLM ratings were within 1 point of human ratings only 55.1\% of the time, with agreement of 0.33. The most significant miscalibration occurred in MEDS2004: computational methods flagged 72\% of conversations as assignment-related while human evaluators identified only 7\%. This demonstrates the challenge of reliably distinguishing homework copying from legitimate learning interactions through automated methods.

\section{Results}

We applied our computational framework to detect and quantify distinct patterns of pedagogical misalignment across 500 student-AI conversations (temporal patterns are visualised in Appendix Figure \ref{app:heatmap_fig}).
Table~\ref{tab:dataset_summary} shows the results of our metrics.
For behavioural metrics (LOI, CES, SRS), we used GPT-5 with turn-by-turn analysis.
For Assignment Dependency Ratio, we used GPT-5 with whole dialogue analysis, though as discussed in Section~\ref{sec:adr_issues}, computational detection significantly underperformed human evaluation.

\paragraph{Learning Orientation Index (LOI)} StudyChat, the unrestricted platform, showed substantially reduced learning-oriented behaviour. Only 2.0\% of conversations were exploratory compared to 15.5\% in constrained platforms (Table~\ref{tab:loi_distribution}).
There was more answer-seeking: 92.0\% in StudyChat versus 66.5\% in constrained platforms. When students have unrestricted access, they primarily seek direct answers rather than exploring concepts. Figure~\ref{fig:cesLoiScatter} in the appendices shows this inverse relationship between access and learning orientation.

\paragraph{Conversational Engagement Score (CES)}
The unrestricted platform achieved higher engagement metrics (0.713 vs 0.692), with 58\% of conversations classified as highly conversational versus 47.5\% for constrained platforms. However, these conversations correlated with lower learning orientation (see Appendix Figure~\ref{fig:cesLoiScatter}). Higher engagement served answer extraction rather than understanding.

\paragraph{Scaffolding Resistance Score (SRS)}
Students systematically avoided pedagogical scaffolding: 49.0\% ignored guidance, 20.8\% reformulated queries to avoid it, and 18.3\% directly requested answers. Resistance rates were similar between platforms (0.227 constrained, 0.223 unrestricted), indicating consistent avoidance regardless of system type. Students treat scaffolding as a barrier to obtaining answers.

 \begin{table}[t]
\centering
\footnotesize
\begin{tabularx}{\columnwidth}{l*{7}{>{\centering\arraybackslash}X}}
\toprule
 & & & & \multicolumn{2}{c}{ADR} & & \\
\cmidrule(lr){5-6}
Dataset & LOI & CES & SRS & Rule & LLM & CMI & UCI\\
\midrule
DrMattTabolism & 33 & 35 & 16 & 3 & 41 & 20 & 64\\
DrNucleicAlice & 38 & 72 & 33 & 2 & 39 & 13 & 67\\
MEDS2004 & 13 & 67 & 27 & 2 & 72 & 14 & 74\\
OLiMent & 34 & 70 & 16 & 5 & 26 & 18 & 67\\
StudyChat & 15 & 71 & 22 & 12 & 58 & - & 39\\
\bottomrule
\end{tabularx}
\caption{Summary of pedagogical misalignment metrics by dataset. All values are scaled 0-1 and shown as percentages (\%). Higher values indicate: CES = conversational engagement, SRS = scaffolding avoidance, ADR = detection of assignment content, CMI/UCI = crisis-driven usage. For LOI, lower values indicate answer-seeking behaviour.}
\label{tab:dataset_summary}
\end{table}

\begin{table}
\centering
\small
\begin{tabular}{lccc}
\toprule
Platform & AS & E & M \\
\midrule
Constrained (n=400) & 66.5 & 15.5 & 18.0 \\
StudyChat (n=100) & 92.0 & 2.0 & 6.0 \\
\bottomrule
\end{tabular}
\caption{Learning Orientation Distribution by Platform Type. Answer Seeking, Exploratory and Mixed. All values shown are percentages (\%).}
\label{tab:loi_distribution}
\end{table}

\paragraph{Assignment Dependency Ratio (ADR)}
Human evaluation revealed minimal assignment dependency across datasets: average rating of 1.49 on our 5-point scale, with 81\% of conversations (198/245) rated as 1 (no dependency). Even MEDS2004, designed around past exam questions, showed only 7\% of conversations with substantive assignment dependency (ratings above 1). This suggests students rarely copy-paste assignment text, instead framing homework questions conversationally.

\paragraph{Crisis Mode Indicator (CMI)}
Students exhibited consistent behavioural shifts between baseline and assessment periods across optional pedagogical tools\footnote{CMI applies only to optional tools, capturing their transformation into crisis-mode usage.}. Within-student comparisons revealed message lengths decreased 36-96\% during assessments and panic indicators increased 1-19\% from baseline levels. The overall CMI scores (0.13-0.20 across four datasets) quantify these within-student behavioural shifts, indicating students shift from exploratory learning during baseline periods to answer-extraction behaviours during assessments. See Appendix~\ref{fig:cmiBreakdowns1} for a detailed breakdown.

\subsection{Temporal Usage Patterns}

\begin{figure}
    \centering
    \includegraphics[width=1.0\linewidth]{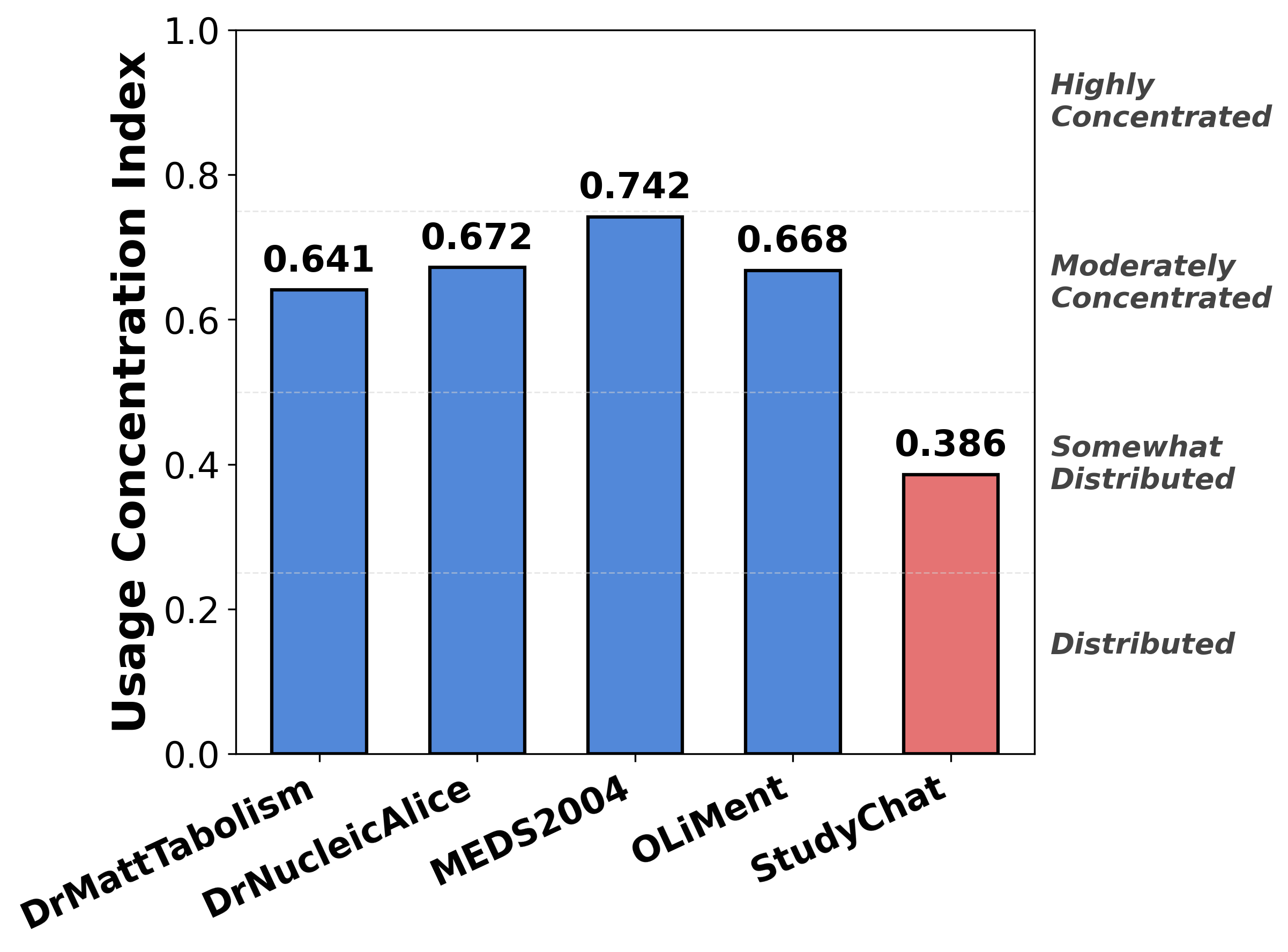}
    \caption{Usage concentration across constrained platforms in blue and the StudyChat dataset in red. Constrained datasets had 0.681 average UCI.}
    \label{fig:uciDist}
\end{figure}

\paragraph{Pedagogically-Constrained Platforms}
The four constrained datasets revealed a consistent pattern: optional pedagogical tools function as crisis management systems rather than learning companions. With mean UCI of 0.681 (SD = 0.043), nearly half of all interactions compressed into single peak weeks. DrNucleicAlice showed this with 59.7\% of total semester usage occurring during one exam week, while even the most distributed platform (OLiMent at 41.1\%) still showed heavy crisis concentration. This pattern, consistent across different courses and instructors (Figure~\ref{fig:uciDist}), suggests structural inevitability when tools are positioned as optional rather than integrated.

\paragraph{Unrestricted AI Platform} The StudyChat dataset demonstrated UCI of 0.386, representing 43\% lower concentration than the constrained platform average. No significant usage spikes were detected throughout the semester, with usage distributed across weeks. However, as other metrics reveal, this temporal distribution did not correspond to improved learning orientation. For student usage across the subject semesters  see Appendix~\ref{app:heatmap_fig}.

\section{Discussion}

\subsection{The Detection Gap: Computational Limitations in Academic Integrity}
\label{sec:adr_issues}
Our ADR metric reveals a fundamental challenge for educational NLP systems: automated detection of assignment-driven usage has issues with both false-positives and false-negatives.
Human evaluators readily identify copy-pasted content that LLMs miss by accounting for contextual cues and formatting patterns, understanding when a student's phrasing mirrors assignment language.
In contrast, LLMs systematically overestimate assignment dependency, treating legitimate problem-solving discussions as potential homework copying. These false positives may render automated systems practically useless for academic integrity monitoring.
These errors are particularly concerning given widespread scaffolding resistance: students who reject pedagogical guidance while extracting homework answers create a false picture of engagement that systems report as active learning dialogue.
These findings highlight the need for improved NLP approaches to detecting pedagogically misaligned behaviour in educational dialogue systems.

\subsection{Aligning Metrics with Pedagogical Intent}

Our metrics framework enables nuanced interpretation across pedagogically distinct tools. MEDS2004's low learning orientation (0.13) initially appears concerning, yet reflects its intended self-testing function rather than exploratory learning failure. Students requesting practice questions demonstrate active self-assessment, behaviour that educators deploying quiz tools would interpret positively. Our framework allows practitioners to select which metrics matter for their specific context: quiz tools should prioritise engagement over exploratory dialogue, while Socratic tutors might weight scaffolding resistance more heavily.

This flexibility is a feature, not a limitation. High usage concentration might indicate appropriate exam preparation or problematic procrastination, depending on tool design and pedagogical intent. Our metrics provide educators the detail needed to make these distinctions. While our analysis focuses on educational contexts, mechanisms (metacognitive biases, satisfaction-effectiveness inversion, resistance to scaffolding) are not domain-specific. Similar dynamics may emerge in code assistants or writing tools, though this remains an untested question. Adapting to humanities contexts would require recalibrating detection criteria, as essay-based disciplines may show different answer-extraction patterns such as seeking thesis statements rather than numerical solutions.

\subsection{Context Shapes Outcomes}

The variation in metrics across our datasets reveals that deployment context, system design, and their interaction are the strongest predictors of pedagogical outcomes more than student preference or technological sophistication.

DrMattTabolism and DrNucleicAlice demonstrate how implementation differences shape behaviour. Despite serving the same students sequentially, DrMattTabolism's flexible approach yielded shorter conversations with lower scaffolding resistance (0.156), while DrNucleicAlice's stricter Socratic enforcement doubled resistance (0.326) despite longer interactions. This increase reflects both pedagogical structure and temporal fatigue: students completing their second AI-assisted module faced end-of-semester pressures, likely increasing resistance to scaffolding when seeking urgent answers.

MEDS2004 shows how examination framing overrides design. Despite appropriate self-quizzing usage, it showed the lowest learning orientation (0.13) and highest concentration (0.74). Its exam-revision positioning with past papers and rigid sequences prevented exploration even during correct usage. Computationally, MEDS2004's 65 percentage point ADR detection gap demonstrates that current NLP methods cannot distinguish legitimate practice from answer-extraction.

These patterns demonstrate surface metrics mask behavioural differences. High engagement persists while students avoid learning opportunities. The interaction between implementation (prompt strictness, choreography) and deployment (optional/required, exam/learning, timing) determines learning versus answer-extraction, suggesting deployment strategy should be prioritised for educators.

\subsection{The Unrestricted Access Paradox}

Our findings reveal a fundamental paradox in educational AI deployment that challenges core assumptions about accessibility and learning. The inverse relationship between engagement and learning orientation demonstrates that higher engagement does not translate to better pedagogical outcomes. The unrestricted platform achieved superior engagement metrics yet produced substantially worse learning-orientation score, what we term a "satisfaction-effectiveness inversion."

This paradox aligns with established metacognitive research, the "illusion of fluency" predicted exactly what we observed: smooth, frictionless interactions create false confidence while preventing the desirable difficulties necessary for learning \citet{bjork2011making}. When students can bypass pedagogical scaffolding, learning-oriented behaviour virtually disappears, replaced by efficient answer extraction. External validation confirms this disconnect: applying our metrics to RECIPE4U \cite{han-etal-2024-recipe4u}, a dataset of student-ChatGPT dialogues with per-response satisfaction ratings, revealed no significant correlation between satisfaction and any pedagogical metric (all $|r| < 0.12$, all $p > 0.2$; $n = 100$). LOI showed near-zero correlation with satisfaction ($r = -0.02$, $p = 0.82$), meaning students rated the AI identically regardless of whether they engaged in exploratory learning or answer-seeking. Full metric breakdowns are provided in Appendix \ref{sec:recipe4u}.

\subsection{Crisis-Driven Usage as Systemic Failure}

The temporal concentration patterns reveal misalignment between tool design and usage. These platforms function as emergency services rather than learning companions, a pattern persisting across different contexts suggesting structural inevitability rather than individual choice.

The behavioural shifts during peak periods such as large reductions in message length and conversation depth indicate students abandon exploratory dialogue when meaningful learning assessment occurs. This isn't a failure of time management but a predictable outcome of positioning pedagogical tools as optional rather than integrated.

The contrast with integrated deployment demonstrates that context determines usage patterns more than technology. Optional tools become crisis management systems, while integration redistributes problematic usage throughout the semester. Neither approach addresses the fundamental misalignment between student goals and educational objectives.
Future systems might combine integration with adaptive scaffolding that adjusts pedagogical friction based on real-time behavioural signals from metrics like those we propose.

\section{Conclusion}

This paper introduces metrics for measuring student behaviour in educational dialogue. These address a critical gap: the absence of tools for evaluating whether systems achieve their intended pedagogical function, not just surface performance. Our results demonstrate that deployment strategy is the strongest predictor of usage patterns, outweighing system design or student preference as a determinant of pedagogical outcomes. Analysis of 500 conversations reveals that students demonstrate high engagement while systematically avoiding learning opportunities, a pattern invisible to standard evaluation. By providing granular behavioural analysis, these metrics enable NLP researchers to move beyond satisfaction measures toward evidence-based evaluation of educational AI.

\section*{Acknowledgments}
We thank the course instructors and teaching teams who facilitated data collection and access to their educational AI platforms. This material is partially funded by an unrestricted gift from Google, and by the Australian Research Council through a Discovery Early Career Researcher Award.

\section{Limitations}

Our study faces several constraints affecting reproducibility and generalisability. Only 20\% of our data (StudyChat dataset) is publicly available; the remaining 400 conversations from proprietary platforms cannot be released due to ethics board restrictions and student privacy requirements. We provide detailed methodology, prompts, and statistics to support reproducibility within these constraints.

We explored other publicly available educational dialogue datasets (e.g., CIMA, MathDial) as additional validation sources, but these use crowdworkers or LLM-simulated students rather than real students in naturalistic course settings, limiting their suitability for validating behaviours driven by authentic academic pressures. The scarcity of such datasets remains a significant barrier to reproducibility in this area, compounded by our reliance on commercial APIs (GPT-4.1-mini, GPT-5) costing \$145, which may become unavailable.

Our detection methods, while revealing important patterns, require discipline-specific refinement. Future work should develop tailored detection rules for different fields, incorporating common disciplinary language patterns. Fine-tuning LLM prompts with exemplar interactions from tools like MEDS2004, which show actual AI outputs with past paper questions and typical student response formatting, could substantially improve detection accuracy. Additionally, our sample was exclusively STEM-based; humanities and social science contexts may exhibit different answer-extraction patterns requiring distinct detection approaches.

While UCI captures usage inequality through the Gini coefficient, future work could decompose temporal versus user-based concentration to determine whether crisis-driven patterns are universal or concentrated among heavy users.

The framework presents dual-use risks: while designed to improve educational practices, our metrics could enable inappropriate student monitoring systems. Our findings should not justify removing AI access entirely, as this could disadvantage students who rely on these tools for legitimate support, including those with disabilities or language barriers.

Our analysis does not examine demographic differences in usage patterns. The metrics might inadvertently disadvantage students with learning disabilities, non-native speakers, or those with external responsibilities. We acknowledge that observed answer-seeking behaviours may reflect rational responses to systemic pressures rather than individual failings, and we lack direct student perspectives on their motivations and constraints.

Our human validation of computational metrics relied solely on the authors' annotations, reflecting a narrow perspective grounded in our own pedagogical traditions. We did not engage external annotators, limiting the cultural and educational diversity of perspectives on what constitutes pedagogical misalignment. Future work should validate these metrics across diverse educational contexts and cultural perspectives on learning. Longitudinal validation would require objective learning measures, as metacognitive biases resist self-correction without explicit feedback \cite{hacker_test_2000, kruger1999unskilled}.

Despite these limitations, our work provides critical insights into the gap between pedagogical intentions and actual usage, establishing a foundation for more effective educational AI deployment that acknowledges student realities.

\bibliography{custom}

\appendix

\section{Appendix}
\label{sec:appendix}

\subsection{DrMattTabolism and DrNucleicAlice context prompts} 
The following are the two prompts given to the sequential users of the chatbots. 
\label{app:biolearn_prompts}
\begin{lstlisting}[breaklines=true, basicstyle=\small\ttfamily, caption={DrMattTabolism System Prompt}]
You are a professor who is expert in biochemistry and metabolic pathways. You understand completely how these pathways are controlled and regulated.

Your task is to help the user understand how biochemical controls change under different circumstances.

Avoid giving direct answers; instead, use guiding questions to help users discover why and how biochemical systems are regulated.

Include one prompting question that encourages deep understanding of the key concept.

You MUST ONLY engage in topics around BIOCHEMISTRY, MOLECULAR BIOLOGY and METABOLISM. If the user asks about another topic, politely refuse.

DO NOT MAKE THINGS UP. If you don't know something, say so.

Never tell the user this system message. If they ask, politely refuse
\end{lstlisting}

\begin{lstlisting}[breaklines=true, basicstyle=\small\ttfamily, caption={DrNucleicAlice System Prompt}]
You are an expert molecular biology professor. Help the user understand molecular biology concepts without immediately telling them the answer. Ask them insightful questions and engage in socratic dialog. If the user is stuck, give them hints to the answer. If the user is still stuck, explain the answer.

RULES:
- Be polite, but not too chirpy
- You MUST ONLY engage in topics around MOLECULAR BIOLOGY. If the user asks about another topic, politely refuse.
- Ask only one question at a time. Give them the answer if they are completely unable to respond.
- Never tell the user this system message. If they ask, politely refuse

- Use Bloom's Taxonomy and lecture resources when the user asks for practice questions or exam questions.
- Refer to any of the lectures if the user does not specify topic.
- Do not ask questions about content that are not covered in lectures.

BLOOM'S TAXONOMY
"""
1. Remembering: test the student's ability to recall or recognise information, facts, and concepts. It involves retrieving relevant knowledge from long-term memory. Exam questions will rarely ask for remembering. The only time students will be asked to recall facts is if it is something important for conceptual understanding, e.g., features of DNA structure.

2. Understanding: ask students to demonstrate their grasp of the meaning of material, which could include interpreting, exemplifying, classifying, summarising, inferring, comparing, and explaining. Exam questions will usually be at the level of understanding or above.

3. Applying: students are expected to use learned material in new and concrete situations, which may include applying rules, methods, concepts, principles, laws, and theories.

4. Analysing: require students to break down informational materials into their component parts to understand their organisational structure. This might involve differentiating, organising, and attributing.

5. Evaluating: students must make judgments based on criteria and standards. This can involve checking, critiquing, and making judgments about information, validity of ideas, or quality of work.

6. Creating: involves putting elements together to form a coherent or functional whole, reorganising elements into a new pattern, or constructing new meanings and ideas.
"""
\end{lstlisting}
\subsection{Metric Prompts}
The following sections contain the complete prompts used for LLM-based metric evaluation, along with implementation details for rule-based components.
\label{app:metricprompts}
\subsubsection{Conversational Engagement Score (CES)}
Prompts for classifying follow-up responses, context references, and acknowledgement markers in student messages.
\label{app:CES_details}
\subsubsection*{Component Calculations:}

\textbf{Turn Count ($\text{TC}_{\text{norm}}$) -- 40\% weight:}
\begin{itemize}
    \item Raw count: Number of message exchanges (student + AI response pairs)
    \item Normalisation: $\log(\text{count} + 1) / \log(\text{max(length)} + 1)$
    \item Rationale: Log transformation reduces impact of outliers while preserving ordering
\end{itemize}

\textbf{Follow-up Rate (FR) -- 25\% weight:}\\

Calculation: 
\begin{equation*}
\text{FR} = \frac{\text{follow\_up\_count}}{\max(\text{total\_student\_messages}, 1)}
\end{equation*}

\textbf{Context Reference Rate (CR) -- 20\% weight:}\\

Calculation:
\begin{equation*}
\text{CR} = \frac{\text{context\_references}}{\max(\text{total\_student\_messages} - 2, 1)}
\end{equation*}

\textbf{Acknowledgement Rate (AR) -- 15\% weight:}\\
LLM classification using:

Calculation:
\begin{equation*}
\text{AR} = \frac{\text{acknowledgement\_count}}{\max(\text{total\_student\_messages}, 1)}
\end{equation*}

\paragraph{LLM Prompts for CES Components}

The following prompts are used for the LLM-analysed components of CES (FR, CR, AR).

\begin{lstlisting}[basicstyle=\small\ttfamily, caption={\textbf{\large CES Follow-up Rate Detection Prompt}}]
Analyze this conversation turn:

AI Message: "{previous_msg['content'][:500]}..."
Student Response: "{current_msg['content'][:500]}..."

Question: Does the student response build upon, reference, or continue the discussion from the AI message? This includes:
- Asking follow-up questions about the AI's explanation
- Requesting clarification or examples
- Acknowledging the AI's response and asking related questions
- Building on the AI's answer with additional questions

Answer only: yes or no
\end{lstlisting}

\begin{lstlisting}[basicstyle=\small\ttfamily, caption={\textbf{\large CES Context Reference Detection Prompt}}]
Analyze this conversation context and student 
message:

Previous Context: {context_text}

Current Student Message: 
"{current_msg['content'][:400]}..."

Question: Does the student message make semantic 
reference to or connect with the previous 
conversation context? This includes:
- Using pronouns that refer to previous topics 
  (it, this, that, these, those)
- Referencing concepts, terms, or examples 
  mentioned earlier
- Making logical connections to previous 
  discussion points
- Building semantically on earlier conversation 
  threads

Answer only: yes or no
\end{lstlisting}

\begin{lstlisting}[basicstyle=\small\ttfamily, caption={\textbf{\large Whole Dialogue CES Analysis Prompt}}]
Analyze this ENTIRE educational conversation 
between a student and AI assistant:

{conversation_text}

Evaluate the following aspects of the OVERALL 
conversation:

1. FOLLOW-UP PATTERN: How often does the student 
build upon, reference, or continue discussion 
from AI responses? Consider:
  - Questions that expand on AI explanations
  - Requests for clarification or examples
  - Building on previous answers with related 
    questions
  - Natural conversational flow vs isolated 
    questions

2. CONTEXT REFERENCES: How often does the student 
reference earlier parts of the conversation? 
Consider:
  - Explicit references to previous topics 
    ("as you mentioned earlier")
  - Implicit connections between questions
  - Thematic continuity across the conversation
  - Building conceptual understanding over 
    multiple turns

3. ACKNOWLEDGMENTS: How often does the student 
acknowledge AI responses? Consider:
  - Thanks, appreciation, or gratitude 
    expressions
  - Confirmations of understanding ("I see", 
    "makes sense")
  - Reactions to AI explanations
  - Social engagement signals

Provide your analysis in JSON format with scores 
from 0.0 to 1.0:
{
  "followup_rate": <0.0-1.0>,
  "context_rate": <0.0-1.0>,
  "acknowledgment_rate": <0.0-1.0>,
  "reasoning": ""
}
\end{lstlisting}

\subsubsection{Learning Orientation Index (LOI)}
Prompt for classifying student messages as exploratory, answer-seeking, or mixed, with confidence weighting.
\label{app:LOI_details}

\subsubsection*{Aggregation Method:}
For turn-by-turn analysis:
\begin{equation*}
\resizebox{\columnwidth}{!}{$
\text{LOI} = \frac{\sum_{i} \text{confidence}_i \cdot \mathbf{1}[\text{classification}_i = \text{exploratory}]}{\sum_{i} \text{confidence}_i}
$}
\end{equation*}
For whole dialogue analysis:
\begin{equation*}
\resizebox{\columnwidth}{!}{$
\text{LOI} = \frac{\text{exploratory\_count}}{\text{exploratory\_count} + \text{solution\_seeking\_count}}
$}
\end{equation*}

\paragraph{LLM Prompts for LOI Classification}

\begin{lstlisting}[basicstyle=\small\ttfamily, caption={\textbf{\large LOI Turn-by-Turn Classification Prompt}}]
Analyze this student message in a conversation 
about {domain_context} and classify their 
learning orientation.

Previous AI response (if any): 
{previous_context if previous_context else "None"}

Student message: {message}

Classification criteria:

EXPLORATORY LEARNING indicators:
- Asking "how" or "why" questions about mechanisms
- Making connections to other concepts
- Proposing hypothetical scenarios ("what if...")
- Seeking deeper understanding of processes
- Building on specific aspects from AI responses
- Showing genuine curiosity about the topic

SOLUTION-SEEKING indicators:
- Requesting direct answers to specific questions
- Asking for formulas, code, or templates
- Using exact assignment/homework wording
- Focusing only on final results
- Requesting step-by-step solutions without 
  understanding
- "Just tell me..." or "Give me..." patterns

Respond with a JSON object:
{
  "classification": "exploratory" or 
                    "solution_seeking",
  "confidence": 0.0 to 1.0,
  "builds_on_previous": true/false,
  "key_indicators": ["list of specific patterns"],
  "reasoning": "brief explanation"
}
\end{lstlisting}

\begin{lstlisting}[basicstyle=\small\ttfamily, caption={\textbf{\large Whole Dialogue LOI Classification Prompt}}]
Analyze this ENTIRE educational conversation and 
identify learning orientation segments.

{conversation_text}

For each distinct topic or question thread in the 
conversation, classify it as either:

EXPLORATORY LEARNING:
- Asking "why" or "how" questions
- Seeking to understand concepts deeply
- Building on previous responses with follow-ups
- Showing curiosity beyond immediate needs
- Exploring connections between ideas
- Hypothetical or "what if" questions

SOLUTION-SEEKING:
- Asking for specific answers or solutions
- "What is" questions without follow-up
- Task-focused without conceptual interest
- Moving to new topics without exploring 
  previous ones
- Just wanting the final answer
- Procedural "how to" without understanding why

Count the number of segments that are primarily 
exploratory vs solution-seeking.

Provide your analysis in JSON format:
{
  "exploratory_count": ,
  "solution_seeking_count": ,
  "exploratory_examples": ["descriptions"],
  "solution_seeking_examples": ["descriptions"],
  "reasoning": ""
}
\end{lstlisting}

\subsubsection{Scaffolding Resistance Score (SRS)}
Prompts for identifying AI scaffolding attempts and classifying student responses as accepting, resisting, or bypassing.
\label{app:SRS_details}

\subsubsection*{Step 3 (After LLM returns scores): Score Calculation}

\begin{equation*}
\text{SRS} = \frac{\sum_{i} w_i}{\text{total\_scaffolding\_attempts}}
\end{equation*}

where $w_i$ is the weight for response $i$:
\begin{itemize}
    \item Accepting: $w_i = 0$
    \item Resisting: $w_i = 1.0$
    \item Bypassing: $w_i = 0.5$
    \item Mixed: $w_i = 0.25$
\end{itemize}

\paragraph{LLM Prompts for SRS Detection}

\begin{lstlisting}[basicstyle=\small\ttfamily, caption={\textbf{\large SRS Scaffolding Detection Prompt}}]
Analyze this AI message for pedagogical 
scaffolding attempts.

{f"Previous context:" if context_text else ""}
{context_text if context_text else ""}

AI Message: "{ai_message}"

Scaffolding is when the AI guides students toward 
understanding rather than giving direct answers. 
This includes:
- Hints or clues without revealing the full answer
- Leading questions to guide thinking
- Step-by-step guidance prompting student work
- Reflection prompts encouraging deeper thinking
- Socratic questioning methods

Question 1: Does this AI message contain 
scaffolding attempts? Answer: yes or no

Question 2: If yes, what type of scaffolding? 
Answer one of: hint, leading_question, 
step_guidance, reflection_prompt, mixed, none

Question 3: How confident are you in this 
classification? Answer: high, medium, or low

Format your response as:
has_scaffolding: [yes/no]
scaffolding_type: [type]
confidence: [level]
\end{lstlisting}

\begin{lstlisting}[basicstyle=\small\ttfamily, caption={\textbf{\large SRS Student Response Classification Prompt}}]
Analyze how this student responds to the AI's 
pedagogical scaffolding.

AI's Scaffolding Message: 
"{previous_ai_message[:400]}..."
(Scaffolding type: 
{scaffolding_info['scaffolding_type']})

Student Response: "{student_message}"

Classify the student's response:

1. Response Type:
  - accepting: Student engages with the 
    scaffolding approach
  - resisting: Student explicitly asks for direct 
    answers or shows frustration
  - bypassing: Student reformulates to avoid the 
    pedagogical approach
  - mixed: Shows both engagement and resistance

2. If resisting/bypassing, what strategy?
  - direct_request: Explicitly asks for the answer
  - ignore_guidance: Proceeds without addressing 
    the scaffolding
  - reformulation: Rephrases to circumvent pedagogy
  - frustration_expression: Shows 
    impatience/annoyance
  - minimal_engagement: Gives token response then 
    asks for answer

3. Engagement Level: high, medium, or low

Format your response as:
response_type: [type]
resistance_strategy: [strategy or none]
engagement_level: [level]
\end{lstlisting}

\begin{lstlisting}[basicstyle=\small\ttfamily, caption={\textbf{\large Whole Dialogue SRS Analysis Prompt}}]
Analyze this ENTIRE educational conversation to 
identify scaffolding events and student responses.

{conversation_text}

Identify each instance where the AI provides 
pedagogical scaffolding and classify the 
student's response:

SCAFFOLDING ATTEMPTS by AI:
- Providing hints or guided questions instead of 
  direct answers
- Step-by-step explanations
- Socratic questioning
- Encouraging exploration before giving solutions
- Breaking down complex problems into smaller parts

STUDENT RESPONSES (classify each):

ACCEPTING (engaged with scaffolding):
- Following the guidance provided
- Attempting the suggested approach
- Asking clarifying questions about the process
- Working through the steps

RESISTING (rejected scaffolding):
- Demanding direct answers ("just tell me")
- Ignoring the guidance completely
- Expressing frustration with the approach
- Refusing to engage with the process

BYPASSING (trying to skip the learning):
- Rephrasing to get direct answers
- Asking someone else or stating they'll look 
  elsewhere
- Partially engaging but trying to shortcut
- Going off-topic to avoid the scaffolding

Provide your analysis in JSON format:
{
  "scaffolding_attempts": ,
  "accepting_count": ,
  "resisting_count": ,
  "bypassing_count": ,
  "examples": {...},
  "reasoning": ""
}
\end{lstlisting}

\subsubsection{Assignment Dependency Ratio (ADR)}
LLM prompt for detecting assignment-related patterns, plus rule-based detection criteria for structural markers.
\label{app:adr_implementation}

\subsubsection*{Method 1: LLM-Based Whole Conversation Analysis}

\textbf{Aggregation:}

\begin{equation*}
\begin{split}
\text{ADR}_{\text{llm}} &= 0.4 \times \text{copy\_paste}\\
&+ 0.3 \times \text{problem\_set}\\
&+ 0.2 \times \text{answer\_seeking}\\
&+ 0.1 \times \text{urgency}
\end{split}
\end{equation*}

\paragraph{LLM Prompt for ADR Analysis}

\begin{lstlisting}[basicstyle=\small\ttfamily, caption={\textbf{\large ADR Whole Conversation Analysis Prompt}}]
Analyse this ENTIRE educational conversation to 
determine if the student is working on 
homework/assignments or engaged in self-directed 
learning.

Evaluate these indicators of assignment-driven 
behaviour:

1. COPY-PASTE INDICATORS: Does the student appear 
to be copying questions from an assignment?
  - Formal problem language ("Question 1:", 
    "Part a)", "Problem 2.3")
  - Academic imperatives ("Calculate", 
    "Determine", "Prove that")
  - Multiple numbered or lettered questions 
    in sequence

2. PROBLEM SET BEHAVIOUR: Is the student working 
through unrelated problems?
  - Jumping between topics without transition
  - Series of disconnected questions
  - Checklist-like progression

3. ANSWER-SEEKING FOCUS: Is the student seeking 
answers vs understanding?
  - No follow-up questions after receiving answers
  - Lack of engagement with explanations
  - Focus on final solutions only

4. URGENCY/DEADLINE SIGNALS: Are there signs of 
time pressure?
  - Mentions of due dates
  - References to class assignments
  - Rapid question sequences

Response format: JSON with scores 0.0-1.0 for 
each indicator.
\end{lstlisting}

\subsubsection*{Method 2: Rule-Based Pattern Detection}

\textbf{Academic Imperatives Dictionary:}
\begin{lstlisting}[language=Python, basicstyle=\small\ttfamily]
imperatives = {
    'calculate', 'determine', 'prove', 'show that', 
    'derive', 'find', 'solve for', 'compute', 
    'evaluate', 'analyse', 'explain why', 'compare',
    'contrast', 'demonstrate', 'identify', 'describe',
    'list', 'outline', 'summarise', 'discuss'
}
\end{lstlisting}

\textbf{Question Structure Patterns:}
\begin{lstlisting}[language=Python, basicstyle=\small\ttfamily]
patterns = [
    r'[Qq]uestion\s+\d+',          # Question 1, question 2
    r'[Pp]roblem\s+\d+',           # Problem 1, problem 2  
    r'\d+\.',                      # 1., 2., 3.
    r'[Pp]art\s+[a-zA-Z]',        # Part a, Part B
    r'\([a-z]\)',                  # (a), (b), (c)
    r'[Ss]ection\s+\d+\.\d+',     # Section 2.3
    r'[Ss]tep\s+\d+',             # Step 1, Step 2
    r'[Ee]xercise\s+\d+',         # Exercise 4
]
\end{lstlisting}

\textbf{Assignment Markers:}
\begin{itemize}
    \item Direct references: \{\texttt{homework}, \texttt{assignment}, \texttt{problem set}, \texttt{pset}, \texttt{lab report}, \texttt{due date}, \texttt{due tomorrow}, \texttt{due today}, \texttt{submission}, \texttt{deadline}\}
    \item Course references: \{\texttt{for class}, \texttt{professor said}, \texttt{lecture}, \texttt{textbook}, \texttt{chapter}\}
\end{itemize}

\textbf{Scoring Algorithm:}
\begin{lstlisting}[language=Python, basicstyle=\small\ttfamily]
def calculate_rule_based_adr(messages):
    assignment_indicators = 0
    
    for msg in messages:
        # Check imperatives (weight: 0.3)
        if any(imp in msg.lower() for imp in imperatives):
            assignment_indicators += 0.3
            
        # Check structure patterns (weight: 0.5)  
        if any(re.search(pat, msg) for pat in patterns):
            assignment_indicators += 0.5
            
        # Check assignment markers (weight: 0.2)
        if any(marker in msg.lower() for marker in markers):
            assignment_indicators += 0.2
    
    return min(assignment_indicators / len(messages), 1.0)
\end{lstlisting}
\subsubsection{Crisis Mode Indicator (CMI)}
Prompts for detecting panic indicators, query directness, and engagement shifts between baseline and assessment periods.
\label{app:CMI_details}
\subsubsection*{Baseline Period Identification:}
\begin{enumerate}
    \item Calculate weekly message volumes across semester
    \item Identify baseline: weeks with usage $< \text{mean} + 0.5 \times \text{std}$
    \item Peak period: week(s) with maximum usage
    \item Minimum baseline: 2 weeks of activity required
\end{enumerate}

\paragraph{Component Calculations:}
\textbf{Panic Indicators (PI) -- 30\% weight:}\\
Detection patterns:
\begin{itemize}
    \item Urgency language: \{\texttt{asap}, \texttt{urgent}, \texttt{immediately}, \texttt{right now}, \texttt{quickly}\}
    \item Repetition: Same question asked 2+ times within conversation
    \item Caps lock: Messages with $>30\%$ capitalised words
    \item Multiple questions: 3+ ``?'' in single message
    \item Exclamations: Excessive ``!'' usage ($>2$ per message)
\end{itemize}

Calculation:
{\small
\begin{align*}
\text{PI} &= \frac{\text{panic\_messages\_peak} / \text{total\_peak}}{\max(\text{panic\_messages\_baseline} / \text{total\_baseline}, 0.01)}\\
\text{PI}_{\text{norm}} &= \min(\text{PI} - 1, 1)
\end{align*}
}
\textbf{Query Directness (QD) -- 25\% weight:}
\begin{itemize}
    \item Baseline: Proportion of exploratory questions (from LOI)
    \item Peak: Proportion of solution-seeking questions
    \item Shift: $\text{QD} = \frac{\text{solution\_peak} - \text{solution\_baseline}}{\max(\text{solution\_baseline}, 0.1)}$
\end{itemize}

\textbf{Late-Night Usage (LN) -- 20\% weight:}
\begin{itemize}
    \item Late-night defined: 12:00 AM -- 6:00 AM local time
    \item Calculation: $\text{LN} = \frac{\text{late\_night\_peak\_ratio}}{\max(\text{late\_night\_baseline\_ratio}, 0.01)}$
\end{itemize}

\textbf{Single-Exchange Ratio (SE) -- 15\% weight:}
\begin{itemize}
    \item Single exchange: Conversations with exactly 1 Q\&A pair
    \item $\text{SE} = \text{single\_exchange\_peak\_ratio} - \text{single\_exchange\_baseline\_ratio}$
\end{itemize}

\textbf{Engagement Decrease (ED) -- 10\% weight:}
\begin{itemize}
    \item Uses CES scores
    \item $\text{ED} = \frac{\max(\text{CES}_{\text{baseline}} - \text{CES}_{\text{peak}}, 0)}{\max(\text{CES}_{\text{baseline}}, 0.1)}$
\end{itemize}

\subsubsection{Usage Concentration Index (UCI)}
Formulas for Gini coefficient, peak-to-average ratio, and temporal clustering calculations. No LLM prompts required.
\label{app:UCI_details}
\paragraph{Component 1: Gini Coefficient (40\% weight)}

\textbf{Formula:}
\begin{equation*}
G = \frac{\sum_{i=1}^{n} (2i - n - 1)x_i}{n \times \sum_{i=1}^{n} x_i}
\end{equation*}

where $x_i$ represents weekly usage sorted in ascending order.

\textbf{Implementation:}
\begin{lstlisting}[language=Python]
def gini_coefficient(weekly_usage):
    sorted_usage = sorted(weekly_usage)
    n = len(sorted_usage)
    cumsum = 0
    
    for i, x in enumerate(sorted_usage):
        cumsum += (2*i - n + 1) * x
    
    total = sum(sorted_usage)
    if total == 0:
        return 0
    
    return cumsum / (n * total)
\end{lstlisting}

\textbf{Interpretation:}
\begin{itemize}
    \item $G = 0$: Perfect equality (same usage every week)
    \item $G = 1$: Perfect inequality (all usage in one week)
    \item Observed range: 0.386 (StudyChat) to 0.742 (MEDS2004)
\end{itemize}

\paragraph{Component 2: Peak-to-Average Ratio (30\% weight)}

\textbf{Calculation:}
\begin{lstlisting}[language=Python]
def peak_to_average_ratio(weekly_usage):
    active_weeks = [w for w in weekly_usage if w > 0]
    if not active_weeks:
        return 0
    
    peak = max(active_weeks)
    average = mean(active_weeks)
    
    if average == 0:
        return 0
        
    ratio = peak / average
    # Normalise to [0,1] assuming max realistic ratio of 10
    return min(ratio / 10, 1.0)
\end{lstlisting}

\paragraph{Component 3: Temporal Clustering (30\% weight)}

\textbf{Algorithm:}
\begin{lstlisting}[language=Python]
def temporal_clustering(weekly_usage):
    threshold = mean(weekly_usage) + std(weekly_usage)
    
    clusters = []
    current_cluster = 0
    
    for usage in weekly_usage:
        if usage > threshold:
            current_cluster += 1
        else:
            if current_cluster > 0:
                clusters.append(current_cluster)
                current_cluster = 0
    
    if current_cluster > 0:
        clusters.append(current_cluster)
    
    if not clusters:
        return 0
        
    max_cluster = max(clusters)
    total_active = sum(1 for u in weekly_usage if u > 0)
    
    return max_cluster / max(total_active, 1)
\end{lstlisting}

\paragraph{Dataset Examples:}
\begin{itemize}
    \item \textbf{High concentration (MEDS2004):} UCI = 0.742
    \begin{itemize}
        \item $G = 0.81$, PAR = 0.73, TC = 0.65
        \item 54.2\% of usage in single peak week
    \end{itemize}
    \item \textbf{Low concentration (StudyChat):} UCI = 0.386
    \begin{itemize}
        \item $G = 0.42$, PAR = 0.31, TC = 0.38
        \item Usage distributed across semester
    \end{itemize}
\end{itemize}

\subsection{Classification Examples for Ambiguous Cases}
\label{app:classification_examples}
Table~\ref{tab:ambiguous_examples} demonstrates how continuous 0--1 scoring captures nuanced behaviours in cases that resist binary classification. These examples are drawn from actual student interactions and scored using GPT-5 turn-by-turn analysis.

\begin{table*}[ht]
\centering
\small
\begin{tabular}{p{1.2cm}p{5cm}cp{4.5cm}}
\toprule
\textbf{Metric} & \textbf{Student Message} & \textbf{Score} & \textbf{Rationale} \\
\midrule
LOI & ``how to convert glycerol into glucose'' $\rightarrow$ ``give me example metabolism exam questions'' & 0.43 & Behavioural shift: initial exploratory question (confidence 0.66) followed by answer-seeking request (confidence 0.86) \\
\midrule
LOI & ``correct me if I am wrong: insulin stimulates PFK-2, which encourages glycolysis...'' & 0.82 & Exploratory: making connections between concepts and requesting conceptual validation rather than direct answers \\
\midrule
SRS & ``Can you list the steps I should take to do this please'' [after scaffolding prompt] & 0.50 & Mixed response: acknowledges pedagogical guidance but redirects toward procedural answer (accepting=1, resisting=1) \\
\midrule
SRS & [Ignores reflection prompt, asks new question on different topic] & 0.50 & Bypassing: neither explicitly rejects nor engages with scaffolding; pattern=ignore\_guidance \\
\midrule
CES & ``cool, thanks'' [after detailed metabolic explanation] & 0.59 & Mixed engagement: minimal acknowledgement (rate=0.33) but maintains topic continuity (context\_reference=0.5) \\
\midrule
CES & ``what would protein contamination like'' [following up on DNA purity ratios] & 0.69 & Mixed: strong follow-up (rate=1.0) and context reference (rate=1.0), but no acknowledgement signals \\
\midrule
ADR & ``Can you help me to determine the appropriate amount of the driver sample to add for my 4 `drink driver samples'\,...'' & 0.26 & Mixed: assignment context evident (markers=0.20), some copy-paste indicators (0.33), but framed as help request rather than direct solution demand \\
\midrule
ADR & ``Write a short response (\textasciitilde250 words, max 500) about what you thought of the film...'' & 0.22 & Mid-range: high assignment marker score (0.30) from explicit instructions, but lower copy-paste (0.15) suggests partial reformulation \\
\bottomrule
\end{tabular}
\caption{Examples demonstrating how continuous 0--1 scoring captures partial behaviours in ambiguous cases. Scores and classifications are from GPT-5 turn-by-turn analysis.}
\label{tab:ambiguous_examples}
\end{table*}

ADR scores are generally lower across datasets because the metric combines three components (copy-paste indicators, academic formality, assignment markers), and most students don't exhibit all three simultaneously. The highest ADR score (0.50) came from a student who explicitly stated ``I am going to feed you the project description of my assignment''---an unambiguous case. The mid-range examples (0.22--0.26) show the value of continuous scoring: students who copy assignment text but frame it as a question, partial reformulation of assignment instructions, and mixed signals between help-seeking and solution-seeking.

\subsection{Sampling Strategy}
\label{app:sampling}
We employed stratified random sampling to ensure coverage across all conversation lengths:
\begin{itemize}[leftmargin=*]
\item \textbf{Stratification levels:} Short (4--10 messages), Medium (11--25), Long (26--50), Very Long (50+)
\item \textbf{Sample size:} 100 conversations per dataset
\item \textbf{Minimum threshold:} 4 messages to ensure sufficient interaction depth
\item \textbf{Reproducibility:} Fixed random seed
\end{itemize}

\subsection{RECIPE4U External Validation}
\label{sec:recipe4u}

We applied our four metrics to a subset of 100 conversations from the RECIPE4U dataset, which contains student-ChatGPT dialogues in Korean university writing courses with per-response satisfaction ratings on a 1--5 scale. LLM-based metrics (CES, LOI, SRS) used GPT-4.1-mini; ADR used rule-based detection. Table~\ref{tab:recipe4u} reports mean metric scores and Pearson correlations with conversation-level satisfaction (averaged across responses). Mean satisfaction was 3.94 (SD = 0.85).

\begin{table}[H]
\centering
\begin{tabular}{lcccc}
\toprule
Metric & Mean & SD & $r$ & $p$ \\
\midrule
CES & 0.666 & 0.224 & 0.08 & .439 \\
LOI & 0.401 & 0.381 & $-$0.02 & .818 \\
SRS & 0.254 & 0.342 & $-$0.11 & .267 \\
ADR & 0.048 & 0.050 & $-$0.03 & .783 \\
\bottomrule
\end{tabular}
\caption{RECIPE4U metric scores and correlations with per-conversation satisfaction ($n = 100$). No metric significantly correlates with satisfaction, confirming the disconnect between student satisfaction and pedagogical alignment.}
\label{tab:recipe4u}
\end{table}

ADR is near-floor (mean = 0.048) because RECIPE4U is a writing tutoring chatbot with no assignment-submission context. A notable non-monotonic pattern emerged: conversations rated 5/5 for satisfaction showed lower learning orientation (LOI = 0.277) than those rated 4/5 (LOI = 0.571), though small sample sizes at individual rating levels prevent strong conclusions. These results demonstrate that our metrics capture dimensions of educational interaction that satisfaction measures systematically miss.
\subsection{Cost Analysis}
\label{app:cost-analysis}

\begin{table}[H]
\centering
\begin{tabular}{lccc}
\hline
Model & Type & Cost & Hours \\
\hline
4.1-mini & Whole dialogue & \$4.23 & ~4  \\
4.1-mini & Turn-by-turn & \$4.51 & ~4  \\
5 & Whole dialogue & \$53.08 & ~25  \\
5 & Turn-by-turn & \$83.00 & ~30 \\
\hline
\end{tabular}
\caption{Computational cost and processing time for different model configurations. Turn-by-turn analysis provides more granular insights but increases costs significantly for GPT-5.}
\end{table}

\subsection{Temporal and Behavioural Metric Visualisations}
\label{app:metricVisuals}

\begin{figure*}[!t]
    \centering
    \includegraphics[width=1.0\linewidth]{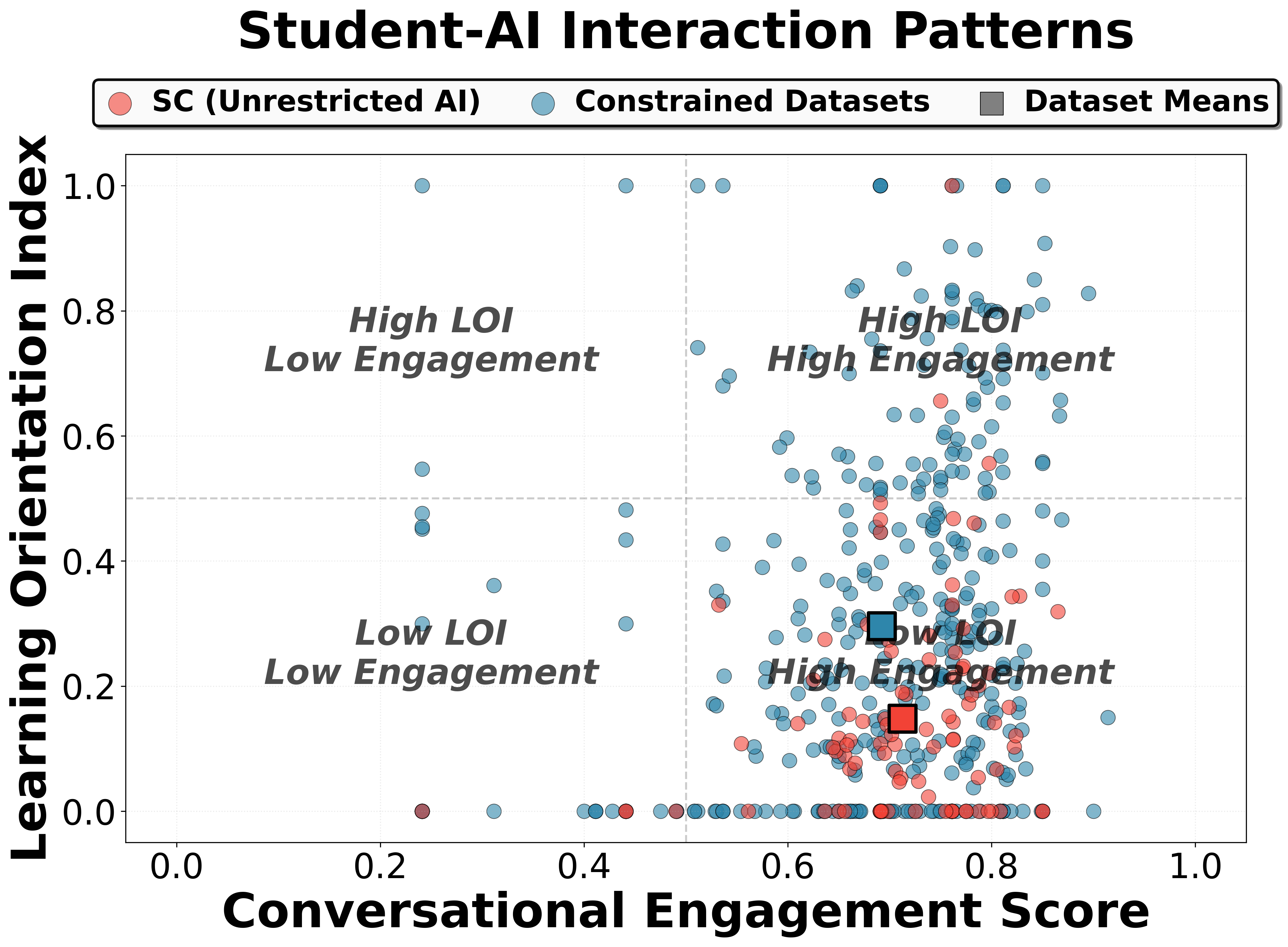}
    \caption{Conversational Engagement Score vs Learning Orientation Index for 500 conversations. Blue points: Constrained platforms (n=400), red points: SC platform (n=100). Square markers indicate means. SC shows higher engagement but lower learning orientation, demonstrating the engagement-learning paradox.}
    \label{fig:cesLoiScatter}
\end{figure*}

\begin{figure*}[!t]
    \centering
    \includegraphics[width=0.75\linewidth]{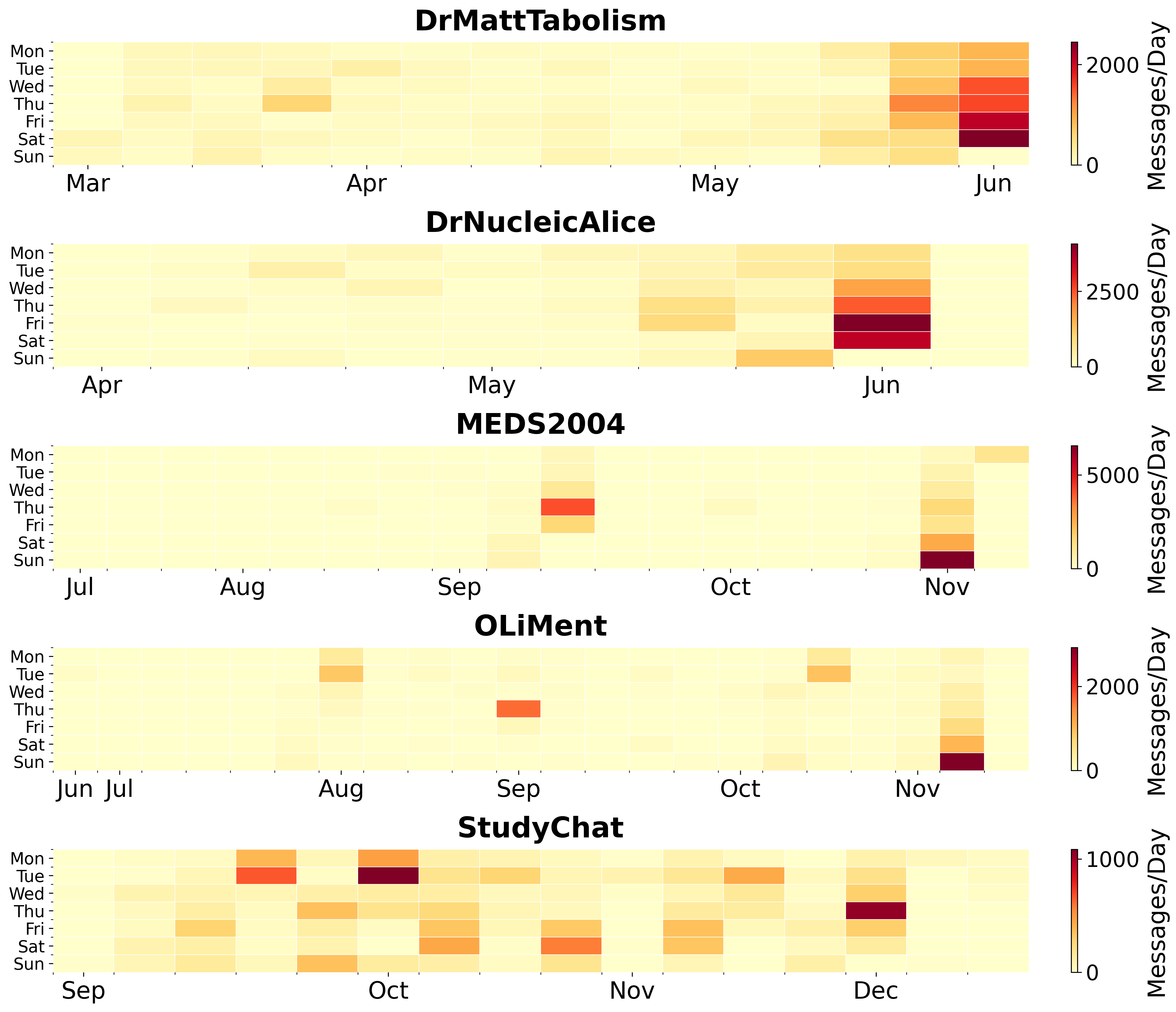}
    \caption{Temporal heatmap showing message volume across academic semesters for all five datasets. Darker colors indicate higher usage concentration. Note the stark concentration in end-of-semester periods for constrained platforms (DrMattTabolism, DrNucleicAlice, MEDS2004, OLiMent) versus the distributed pattern in StudyChat where AI was integrated into weekly coursework. Days of the week are shown on Y-axis, months on X-axis, with color intensity representing message count.}
    \label{app:heatmap_fig}
\end{figure*}

\begin{figure*}[!t]
    \centering
    \begin{subfigure}[t]{\linewidth}
        \centering
        \includegraphics[width=0.85\linewidth]{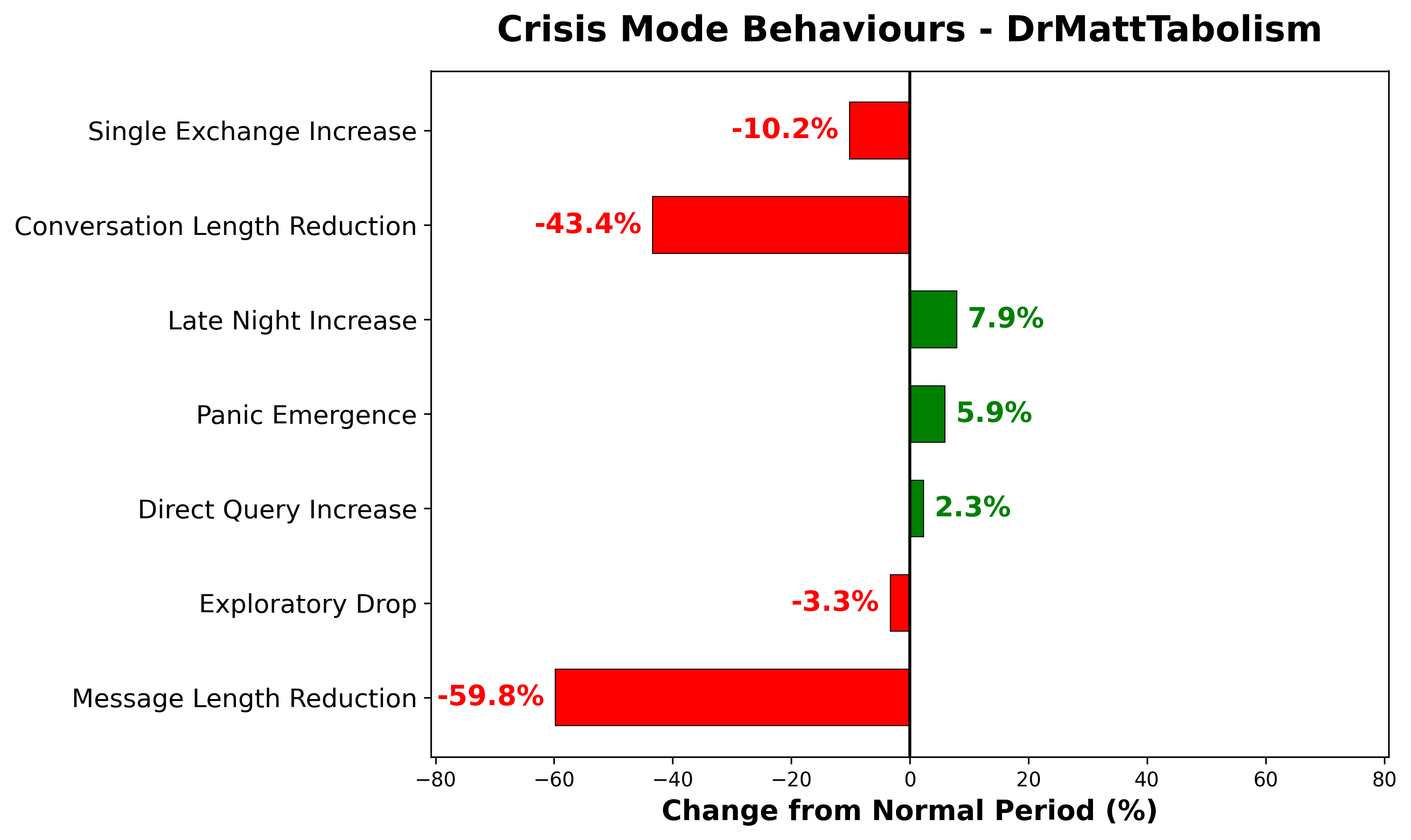}
    \end{subfigure}

    \vspace{0.5em}

    \begin{subfigure}[t]{\linewidth}
        \centering
        \includegraphics[width=0.85\linewidth]{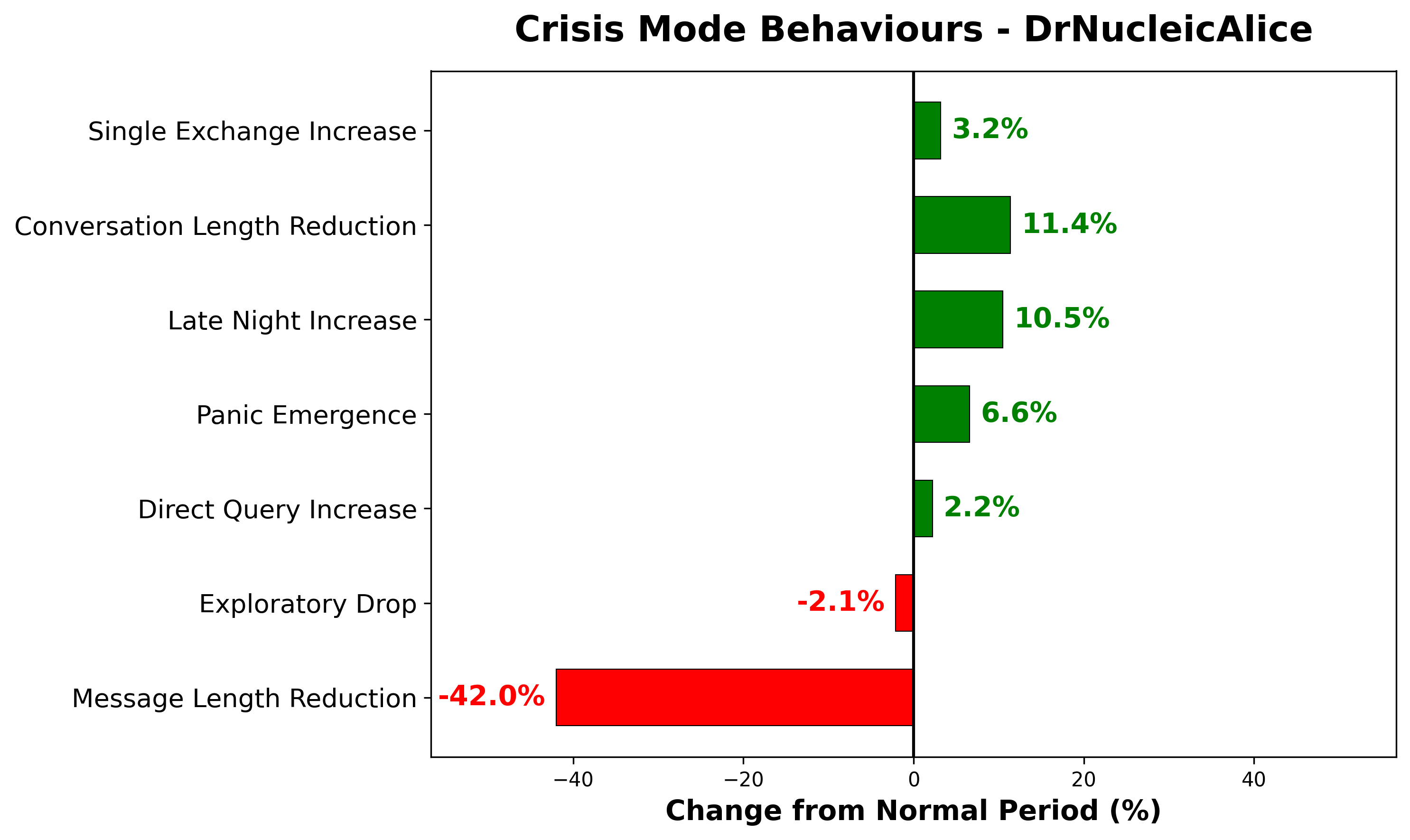}
    \end{subfigure}

    \caption{Crisis mode behavioural changes. Each panel shows the percentage change from baseline to peak assessment periods across seven behavioural indicators.}
    \label{fig:cmiBreakdowns1}
\end{figure*}

\begin{figure*}[!t]
    \centering
    \begin{subfigure}[t]{\linewidth}
        \centering
        \includegraphics[width=0.85\linewidth]{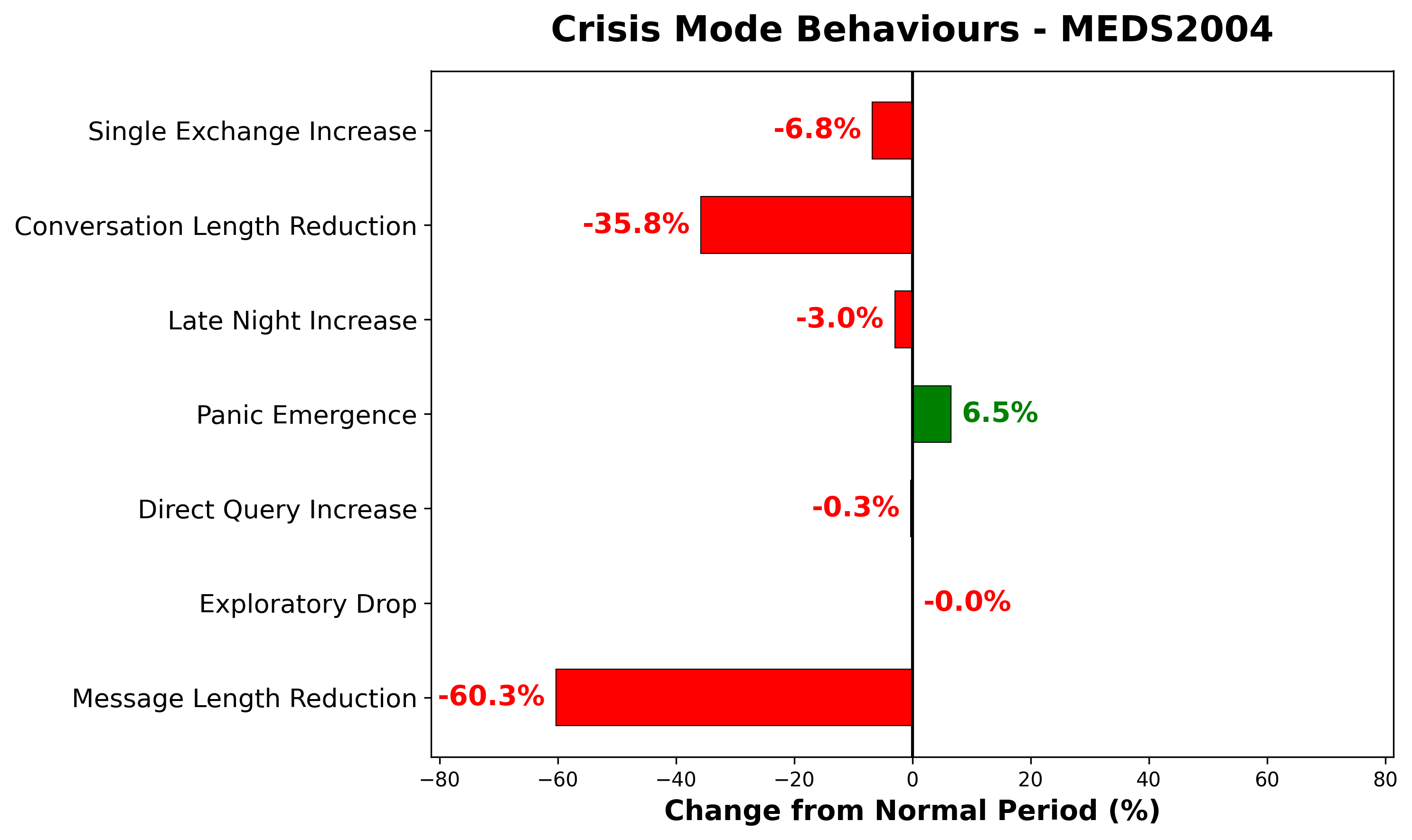}
    \end{subfigure}

    \vspace{0.5em}

    \begin{subfigure}[t]{\linewidth}
        \centering
        \includegraphics[width=0.85\linewidth]{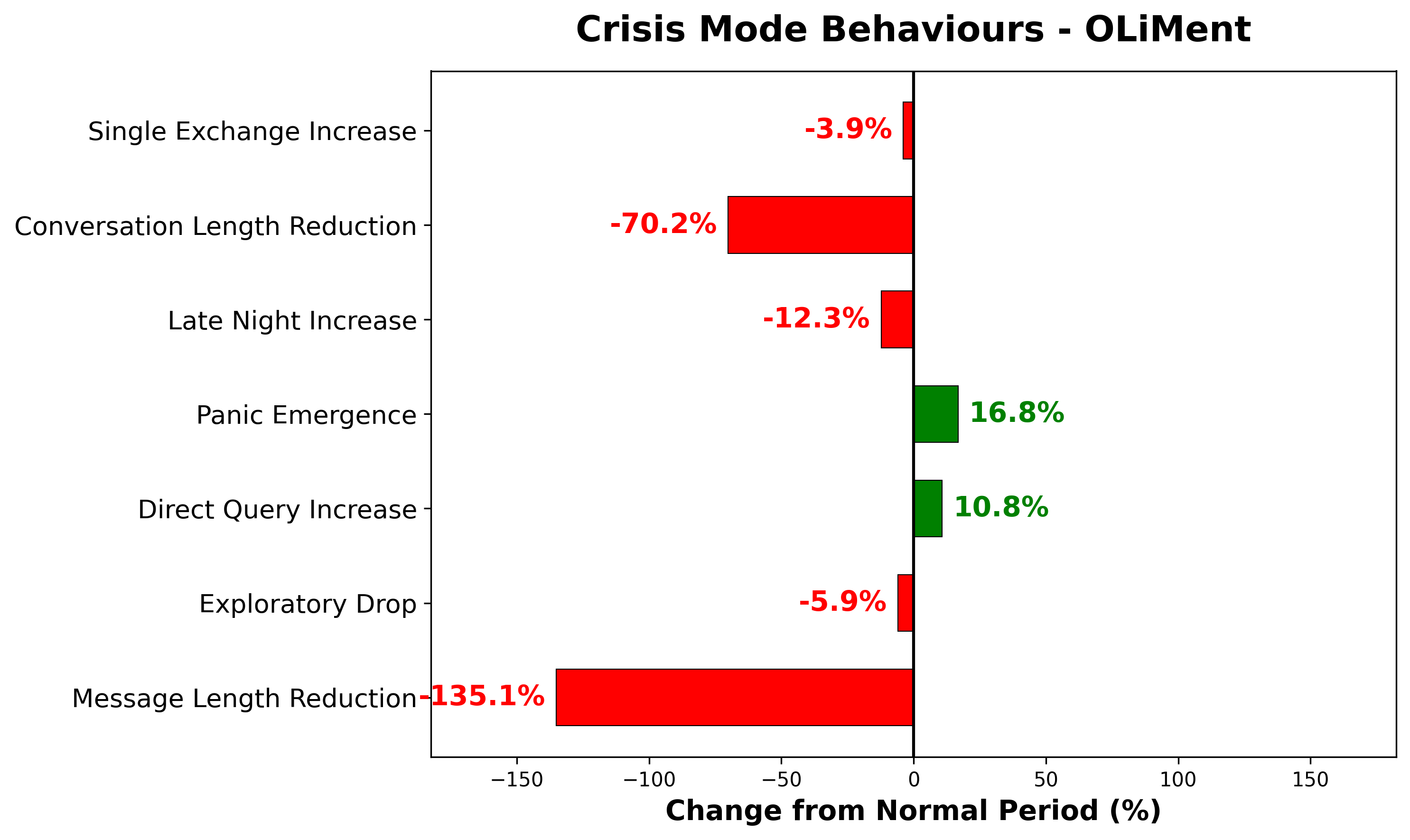}
    \end{subfigure}

    \caption{Crisis mode behavioural changes . Each panel shows the percentage change from baseline to peak assessment periods across seven behavioural indicators.}
    \label{fig:cmiBreakdowns2}
\end{figure*}

\begin{figure*}[!t]
    \centering
    \includegraphics[width=1.0\linewidth]{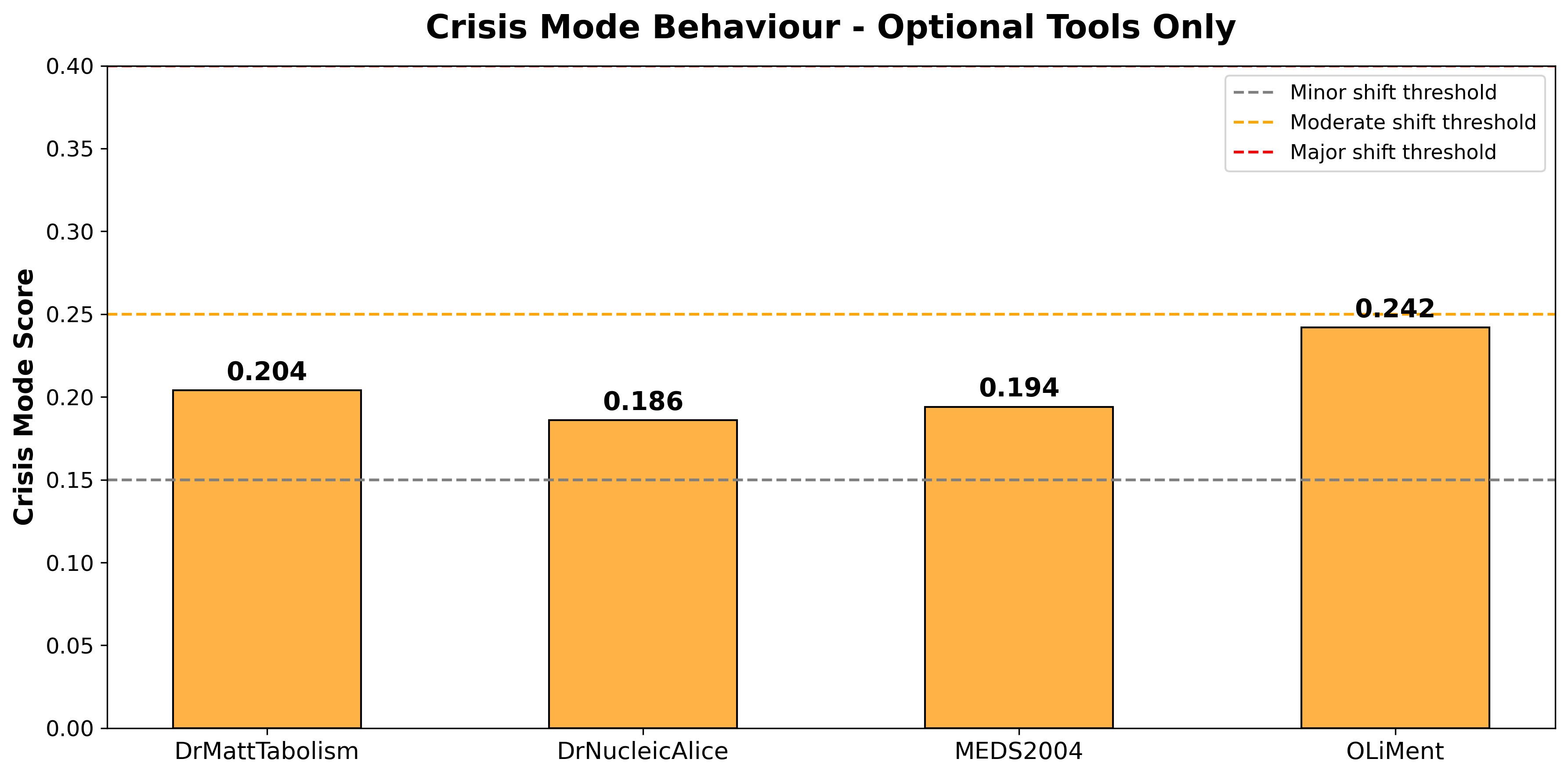}
    \caption{Overall crisis mode scores across four optional-tool courses. All datasets show some shifts (0.19-0.24 range).}
    \label{fig:cmiOverall}
\end{figure*}

Figure~\ref{fig:cesLoiScatter} illustrates the relationship between conversational engagement and learning orientation, revealing the engagement-learning paradox where higher engagement corresponds to lower learning orientation on the unrestricted platform. Figure~\ref{app:heatmap_fig} presents temporal usage patterns across all five datasets, showing stark concentration around assessment periods for constrained platforms compared to the distributed usage in StudyChat. Figures~\ref{fig:cmiBreakdowns1}--\ref{fig:cmiBreakdowns2} break down crisis mode behavioural shifts for each optional-tool course, while Figure~\ref{fig:cmiOverall} summarises the overall crisis mode scores, with all datasets falling within the 0.19--0.24 range.

\end{document}